\documentclass[10pt,twocolumn,letterpaper]{article}

\usepackage{iccv}
\usepackage{times}
\usepackage{epsfig}
\usepackage{graphicx}
\usepackage{amsmath}
\usepackage{amssymb}

\usepackage{booktabs}
\usepackage{siunitx}
\usepackage{multirow}
\usepackage{algorithm}
\usepackage{algorithmicx}
\usepackage{algpseudocode}
\usepackage{bbm}

\usepackage[pagebackref=true,breaklinks=true,colorlinks,bookmarks=false]{hyperref}

\iccvfinalcopy %

\def\iccvPaperID{904} %

\ificcvfinal\fi
\begin{document}
\title{Universally Slimmable Networks and Improved Training Techniques}

\author{
  Jiahui Yu
  \qquad
  Thomas Huang\\
  University of Illinois at Urbana-Champaign
}

\maketitle
\ificcvfinal\thispagestyle{empty}\fi

\begin{abstract}
Slimmable networks~\cite{yu2018slimmable} are a family of neural networks that can instantly adjust the runtime width. The width can be chosen from a predefined widths set to adaptively optimize accuracy-efficiency trade-offs at runtime. In this work, we propose a systematic approach to train universally slimmable networks (US-Nets), extending slimmable networks to execute at arbitrary width, and generalizing to networks both with and without batch normalization layers. We further propose two improved training techniques for US-Nets, named the sandwich rule and inplace distillation, to enhance training process and boost testing accuracy. We show improved performance of universally slimmable MobileNet v1 and MobileNet v2 on ImageNet classification task, compared with individually trained ones and 4-switch slimmable network baselines. We also evaluate the proposed US-Nets and improved training techniques on tasks of image super-resolution and deep reinforcement learning. Extensive ablation experiments on these representative tasks demonstrate the effectiveness of our proposed methods. Our discovery opens up the possibility to directly evaluate FLOPs-Accuracy spectrum of network architectures.
Code and models are available at: \url{https://github.com/JiahuiYu/slimmable_networks}.
\end{abstract}

\section{Introduction}
\begin{figure}[ht]
\centering
\includegraphics[width=\linewidth]{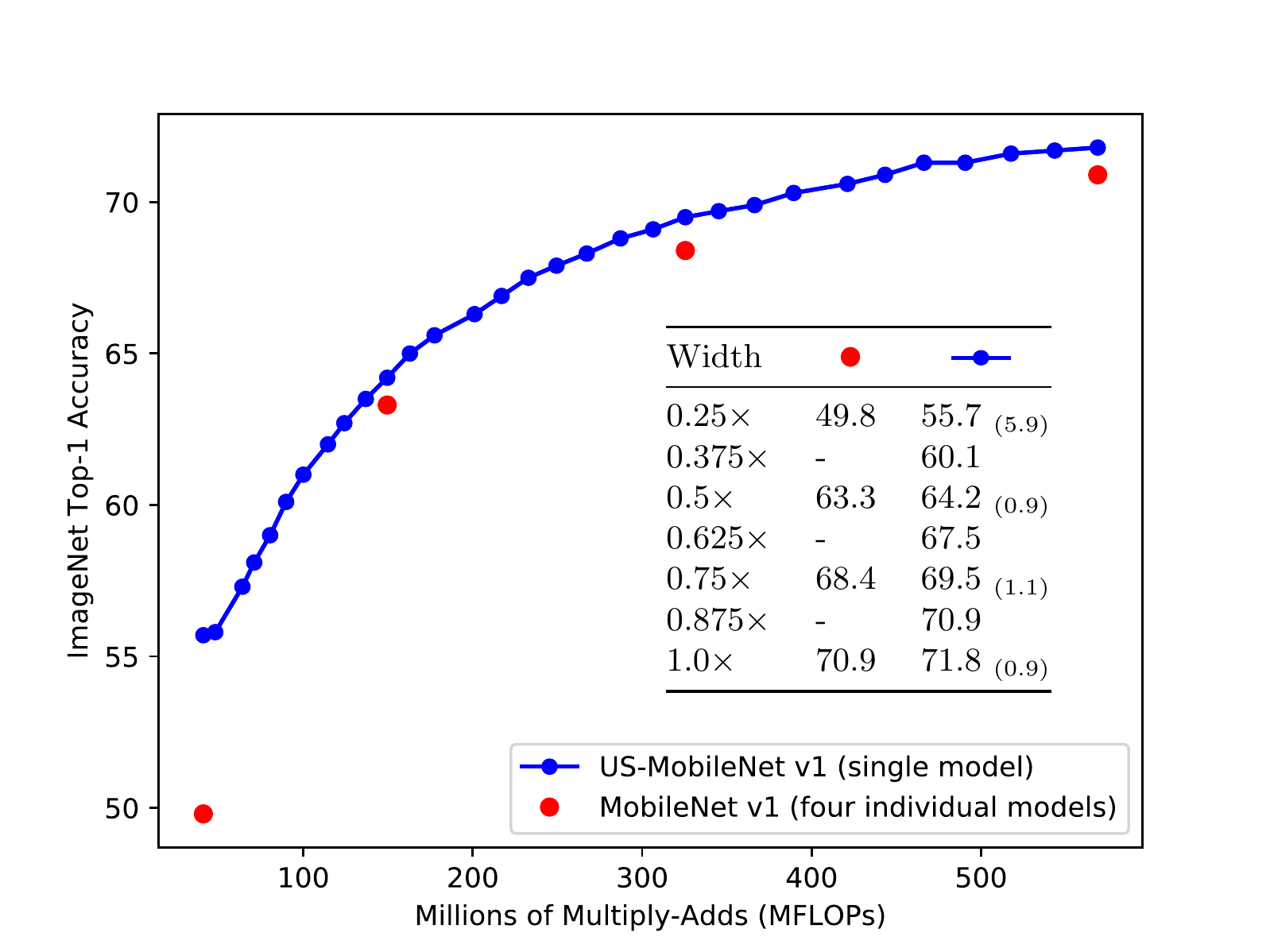}
\caption{FLOPs-Accuracy spectrum of \textbf{single} US-MobileNet v1 model, compared with \textbf{four individual} MobileNet v1 models.}
\label{figs:teaser}
\end{figure}

The ability to run neural network models within latency budget is of paramount importance for applications on mobile phones, augmented reality glasses, self-driving cars, security cameras and many others~\cite{yang2018efficient, huang2018data, ren2015faster}. Among these applications, many are required to deploy trained models across different devices or hardware versions~\cite{yu2018slimmable, huang2017multi, liu2017dynamic}. However, a single trained network cannot achieve optimal accuracy-efficiency trade-offs across different devices (\eg, face recognition model running on diverse mobile phones). To address the problem, recently slimmable networks~\cite{yu2018slimmable} were introduced that can switch among different widths at runtime, permitting instant and adaptive accuracy-efficiency trade-offs. The width can be chosen from a predefined widths set, for example {\([0.25, 0.5, 0.75, 1.0]\times\)}, where  {\([\cdot]\times\)} denotes available widths, and {\(0.25 \times\)} represents that the width in all layers is scaled by \(0.25\) of the full model. To train a slimmable network, switchable batch normalization~\cite{yu2018slimmable} is proposed that privatizes batch normalization~\cite{ioffe2015batch} layers for each sub-network. A slimmable network has accuracy similar to that of individually trained ones with the same width~\cite{yu2018slimmable}.

Driven by slimmable networks, a further question arises: \textit{can a single neural network run at arbitrary width}? The question motivates us to rethink the basic form of feature aggregation. In deep neural networks, the value of a single output neuron is an aggregation of all input neurons weighted by learnable coefficients \(y = \sum_{i=1}^{n} w_{i} x_{i}\), where \(x\) is input neuron, \(y\) is output neuron, \(w\) is learnable coefficient and \(n\) is number of input channels. This formulation indicates that each input channel or group of channels can be viewed as a \textit{residual component}~\cite{he2016deep} to an output neuron. Thus, a wider network should have no worse performance than its slim one (the accuracy of slim one can always be achieved by learning new connections to zeros). In other words, if we consider a single layer, the residual error between full aggregation and partial aggregation decreases and is bounded:
\begin{equation} \label{eqs:intro}
|y^{n} - y^{k+1}| \leq |y^{n} - y^{k}| \leq |y^{n} - y^{k_0}|,
\end{equation}
where \(y^{k}\) summarizes the first \(k\) channels \(y^{k} = \sum_{i=1}^{k} w_{i} x_{i}\), \(\forall k \in [k_0, n)\), \(k_0\) is a constant hyper-parameter (for example, \(k_0 = \lceil0.25n\rceil\)).
The bounded inequality\footnote{The analysis is based on a single hidden layer. Future research on theoretical analysis of deep neural networks with nonlinear activation may fully reveal why or why not universally slimmable networks exist.} suggests that a slimmable network~\cite{yu2018slimmable} executable at a discrete widths set can potentially run at any width in between (if properly trained), since the residual error decreases by the increase of width and is bounded. Moreover, the inequality conceptually applies to any deep neural network, regardless of what normalization layers~\cite{ioffe2015batch, salimans2016weight} are used. However, as suggested in~\cite{yu2018slimmable}, batch normalization (BN)~\cite{ioffe2015batch} requires special treatment because of the inconsistency between training and testing. 

In this work, we present universally slimmable networks (US-Nets) that can run at any width in a wide range. Three fundamental challenges of training US-Nets are addressed. First, how to deal with neural networks with batch normalization? Second, how to train US-Nets efficiently? Third, compared with training individual networks, what else can we explore in US-Nets to improve overall performance?

Batch normalization~\cite{ioffe2015batch} has been one of the most important components in deep learning. During training, it normalizes feature with mean and variance of current mini-batch, while in inference, moving averaged statistics of training are used instead. This inconsistency leads to failure of training slimmable networks, as shown in~\cite{yu2018slimmable}. The switchable batch normalization~\cite{yu2018slimmable} (we address the version of shared scale and bias by default, the version of private scale and bias will be discussed in Section~\ref{secs:discuss}) is then introduced. However, it is not practical for training US-Nets for two reasons. First, accumulating independent BN statistics of all sub-networks in a US-Net during training is computationally intensive and inefficient. Second, if in each iteration we only update some sampled sub-networks, then these BN statistics are insufficiently accumulated thus inaccurate, leading to much worse accuracy in our experiments. To properly address the problem, we adapt the batch normalization with a simple modification. The modification is to calculate BN statistics of all widths after training. The weights of US-Nets are fixed after training, thus all BN statistics can be computed in parallel on cluster servers. More importantly, we find that a \textit{randomly sampled subset} of training images, as few as \num{1} mini-batch (\num{1024} images), already produces accurate estimation. Thus calculating BN post-statistics can be very fast. We note that to be more general, we intentionally avoid modifying the formulation of BN or proposing new normalization.

Next we propose an improved training algorithm for US-Nets motivated by the bounded inequality in Equation~\ref{eqs:intro}. To train a US-Net, a natural solution is to accumulate or average losses sampled from different widths. For example, in each training iteration we randomly sample \(n\) widths in the range of {
\([0.25, 1.0]\times\)}. Taking a step further, we should notice that in a US-Net, performances at all widths are bounded by performance of the model at smallest width (\eg, {\(0.25\times\)}) and largest width (\eg, {
\(1.0\times\)}). In other words, optimizing performance lower bound and upper bound can implicitly optimize the model at all widths. Thus, instead of sampling \(n\) widths randomly, in each training iteration we train the model at smallest width, largest width and (\(n\)-\(2\)) randomly sampled widths. We employ this rule (named \textit{the sandwich rule}) to train US-Nets and show better convergence behavior and overall performance.

Further we propose \textit{inplace distillation} that transfers knowledge inside a single US-Net from full-network to sub-networks \textit{inplace} in each training iteration. The idea is motivated by two-step knowledge distilling~\cite{hinton2015distilling} where a large model is trained first, then its learned knowledge is transferred to a small model by training with predicted soft-targets. In US-Nets, by \textit{the sandwich rule} we train the model at largest width, smallest width and other randomly sampled widths all together in each iteration. Remarkably, this training scheme naturally supports inplace knowledge transferring: we can directly use the predicted label of the model at the largest width as the training label for other widths, while for the largest width we use ground truth. It can be implemented inplace in training without additional computation and memory cost. Importantly, the proposed \textit{inplace distillation} is general and we find it works well not only for image classification, but also on tasks of image super-resolution and deep reinforcement learning.

\begin{figure*}[ht]
\centering
\includegraphics[width=\linewidth]{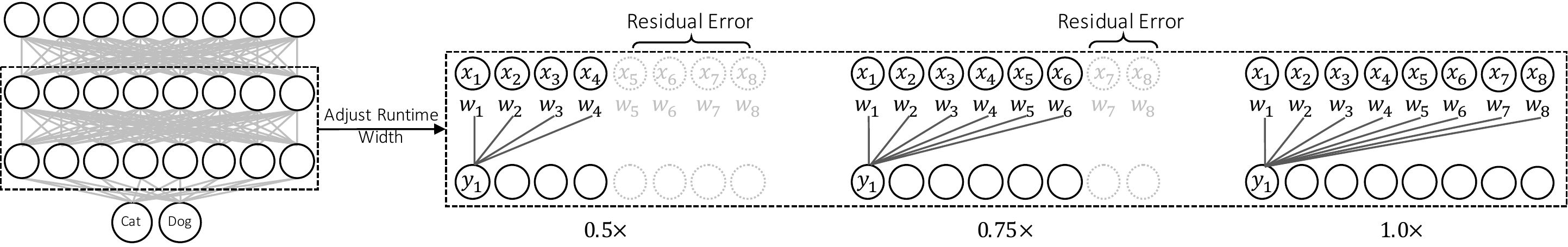}
\caption{Illustration of a network executing at different widths. We specifically consider an output neuron \(y_1\) in a layer (right, zoomed).}
\label{figs:rethink_neuron}
\end{figure*}

We apply the proposed methods to train universally slimmable networks on representative tasks with representative networks (both with and without BN, and both residual and non-residual networks). We show that trained US-Nets perform similarly or even better than individually trained models. Extensive ablation studies on \textit{the sandwich rule} and \textit{inplace distillation} demonstrate the effectiveness of our proposed methods. Our contributions are summarized as follows:
\begin{enumerate}
    \setlength\itemsep{0.05em}
	\item For the first time we are able to train a single neural network executable at arbitrary width, using a simple and general approach.
	\item We further propose two improved training techniques in the context of US-Nets to enhance training process and boost testing accuracy.
	\item We present experiments and ablation studies on image classification, image super-resolution and deep reinforcement learning. 
	\item We further intensively study the US-Nets with regard to (1) width lower bound \(k_0\), (2) width divisor \(d\), (3) number of sampled widths per training iteration \(n\), and (4) size of subset for BN post-statistics \(s\).
	\item We further show that our method can also be applied to train nonuniform US-Nets where each layer can adjust its own width ratio, instead of a global width ratio uniformly applied on all layers.
	\item Our discovery opens up the possibility to many related fields, for examples, network comparison in terms of FLOPs-Accuracy spectrum (Figure~\ref{figs:teaser}), and one-shot architecture search for number of channels~\cite{yu2019network}.
\end{enumerate}

\section{Related Work}
\textbf{Slimmable Networks.} Yu \etal~\cite{yu2018slimmable} present the initial approach to train a single neural network executable at different widths, permitting instant and adaptive accuracy-efficiency trade-offs at runtime. The width can be chosen from a predefined widths set. The major obstacle of training slimmable networks is addressed: accumulating different numbers of channels results in different feature mean and variance. This discrepancy across different sub-networks leads to inaccurate statistics of shared Batch Normalization layers~\cite{ioffe2015batch}. Switchable batch normalization is proposed that employs independent batch normalization for different sub-networks in a slimmable network. On tasks of image recognition (\ie, classification, detection and segmentation), slimmable networks achieve accuracy similar to that of individually trained models~\cite{yu2018slimmable}.

\textbf{Knowledge Distilling.} The idea of knowledge distilling~\cite{hinton2015distilling} is to transfer the learned knowledge from a pretrained network to a new one by training it with predicted features, soft-targets or both. It has many applications in computer vision, network compression, reinforcement learning and sequence learning problems~\cite{bagherinezhad2018label, chen2015net2net, kim2016sequence, parisotto2015actor, romero2014fitnets}. FitNet~\cite{romero2014fitnets} proposes to train a thinner network using both outputs and intermediate representations learned by the teacher network as hints. Net2Net~\cite{chen2015net2net} proposes to transfer the knowledge from a pretrained network to new deeper or wider one for accelerating training. Actor-Mimic~\cite{parisotto2015actor} trains a single policy network to behave in multiple tasks with guidance of many teacher networks. Knowledge distillation is also effectively applied to word-level prediction for neural machine translation~\cite{kim2016sequence}.

\section{Universally Slimmable Networks}

\subsection{Rethinking Feature Aggregation} \label{secs:rethink}
Deep neural networks are composed of layers where each layer is made of neurons. As the fundamental element of deep learning, a neuron performs weighted sum of all input neurons as its value, propagating layer by layer to make final predictions. An example is shown in Figure~\ref{figs:rethink_neuron}. The output neuron \(y\) is computed as:
\begin{equation}
    y = \sum_{i=1}^{n} w_{i} x_{i},
\end{equation}
where \(n\) is the number of input neurons (or channels in convolutional networks), \(x = \{x_1, x_2, ..., x_n\}\) is input neurons, \(w = \{w_1, w_2, ..., w_n\}\) is learnable coefficient, \(y\) is a single output neuron. This process is also known as \textit{feature aggregation}: each input neuron is responsible for detecting a particular feature, and the output neuron aggregates all input features with learnable transformations.

The number of channels in a network is usually a manually picked hyper-parameter (\eg, \num{128}, \num{256}, ..., \num{2048}). It plays a significant role in the accuracy and efficiency of deep models: wider networks normally have better accuracy with sacrifice of runtime efficiency. To provide the flexibility, many architecture engineering works~\cite{howard2017mobilenets, sandler2018inverted, zhang2017shufflenet} individually train their proposed networks with different width multipliers: a global hyper-parameter to slim a network uniformly at each layer.

We aim to train a single network that can directly run at arbitrary width. It motivates us to rethink the basic form of feature aggregation in deep neural networks. As shown in Figure~\ref{figs:rethink_neuron}, feature aggregation can be explicitly interpreted in the framework of \textit{channel-wise residual learning}~\cite{he2016deep}, where each input channel or group of channels can be viewed as a \textit{residual component}~\cite{he2016deep} for the output neuron. Thus, a wider network should have no worse performance than its slim one (the accuracy of slim one can always be achieved by learning new connections to zeros). In other words, the residual error \(\delta\) between fully aggregated feature \(y^n\) and partially aggregated feature \(y^k\) decreases and is bounded:
\begin{equation} \label{eqs:rethink}
 0 \leq \delta_{k+1} \leq \delta_{k} \leq \delta_{k_0}, \delta_k = |y^{n} - y^{k}|,
\end{equation}
where \(y^{k}\) summarizes the first \(k\) channels \(y^{k} = \sum_{i=1}^{k} w_{i} x_{i}\), \(\forall k \in [k_0, n)\), \(k_0\) is a constant hyper-parameter (for example, \(k_0 = \lceil0.25n\rceil\)).

The bounded inequality in Equation~\ref{eqs:rethink} provides clues about several speculations: (1) Slimmable network~\cite{yu2018slimmable} executable at a discrete widths set can potentially run at any width in between (if properly trained). In other words, a single neural network may execute at any width in a wide range for \(k\) from \(k_0\) to \(n\), since the residual error of each feature is bounded by \(\delta_{k_0}\), and decreases by increase of width \(k\). (2) Conceptually the bounded inequality applies to any deep neural network, regardless of what normalization layers (\eg, batch normalization~\cite{ioffe2015batch} and weight normalization~\cite{salimans2016weight}) are used. 
Thus, in the following sections we mainly explore how to train a single neural network executable at arbitrary width. These networks are named as \textit{universally slimmable networks}, or simply \textit{US-Nets}.

\subsection{Post-Statistics of Batch Normalization} \label{secs:post_statistics}
However, as suggested in~\cite{yu2018slimmable}, batch normalization~\cite{ioffe2015batch} requires special treatment because of the inconsistency between training and testing. During training, features in each layer are normalized with mean and variance of the current mini-batch feature values \(x_B\):
\begin{equation}
    \hat x_B = \gamma \frac{x_B - E_B[x_B]}{\sqrt{Var_B[x_B]+\epsilon}} + \beta,
\end{equation}
where \(\epsilon\) is a small value (e.g.\ \(10^{-5}\)) to avoid zero-division, \(\gamma\) and \(\beta\) are learnable scale and bias. The values of feature mean and variance are then updated to global statistics as moving averages:
\begin{equation} \label{eqs:moving_averages}
    \begin{split}
    \mu_t &= m \mu_{t-1} + (1-m) E_B[x_B],\\
    \sigma^2_{t} &= m \sigma^2_{t-1} + (1-m) Var_B[x_B], 
    \end{split}
\end{equation}
where \(m\) is the momentum (\eg, \(0.9\)), and \(t\) is the index of training iteration. We denote \(\mu = \mu_T, \sigma^2 = \sigma^2_T\), assuming the network is trained for \(T\) iterations totally. During inference, these global statistics are used instead:
\begin{equation} \label{eqs:bn_test}
\hat x_{\mathit{test}} = \gamma^* \frac{x_{\mathit{test}} - \mu}{\sqrt{\sigma^2+\epsilon}} + \beta^*,
\end{equation}
where \(\gamma^*\) and \(\beta^*\) are the optimized scale and bias. Note that after training, the Equation~\ref{eqs:bn_test} can be reformulated as a simple linear transformation:
\begin{equation} \label{eqs:merge}
    \hat x_{\mathit{test}} =  \gamma'x_{\mathit{test}} + \beta', \gamma' = \frac{\gamma^*}{\sqrt{\sigma^2 + \epsilon}}, \beta' = \beta^* - \gamma'\mu,
\end{equation}
and usually \(\gamma'\) and \(\beta'\) can be further fused into its previous convolution layer.

In slimmable networks, accumulating different numbers of channels results in different feature means and variances, which further leads to inaccurate statistics of shared BN~\cite{yu2018slimmable}. Yu \etal introduced switchable batch normalization that privatizes \(\gamma\), \(\beta\), \(\mu\), \(\sigma^2\) of BN for each sub-network. Although parameter \(\gamma\), \(\beta\) can be merged after training (Equation~\ref{eqs:merge}), slimmable networks with shared \(\gamma\) and \(\beta\) have close performance~\cite{yu2018slimmable}.

Regarding universally slimmable networks, however, switchable batch normalization~\cite{yu2018slimmable} is not practical for two reasons. First, accumulating independent BN statistics of all sub-networks in a US-Net during training is computationally intensive and inefficient. For example, assuming an \(n\mathit{-channel}\) layer can adjust its width from \(\lceil0.25n\rceil\) to \(n\), totally there are (\(n - \lceil0.25n\rceil\)) sub-networks to evaluate and \(\lceil0.25n\rceil + (\lceil0.25n\rceil+1) + ... + n = \mathcal{O}(n^2)\) variables of BN statistics to update in each training iteration. Second, if in each iteration we only update some sampled sub-networks, then these BN statistics are insufficiently accumulated thus inaccurate, leading to much worse accuracy in our experiments.

To this end, we adapt the batch normalization with a simple modification that can properly address the problem. The modification is to calculate BN statistics of all widths after training. Trainable parameters of US-Nets are fixed, thus all BN statistics can be computed in parallel on cluster servers. After training, we can calculate BN statistics over training samples, either as moving averages in Equation~\ref{eqs:moving_averages} or exact averages as follows:
\begin{equation} \label{eqs:exact_averages}
    \begin{split}
    m &= (t-1)/t,\\
    \mu_t &= m \mu_{t-1} + (1-m) E_B[x_B],\\
    \sigma^2_{t} &= m \sigma^2_{t-1} + (1-m) Var_B[x_B].
    \end{split}
\end{equation}
Our experiments show that exact averages have slightly better performance than moving averages.

In practice, we find it is not necessary to accumulate BN statistics over all training samples: a randomly sampled subset (\eg, \(1k\) images) already produces accurate estimations. With this option, calculating post-statistics of BN can be extremely fast (by default we calculate over all training samples). In experiments, we will compare the accuracy for different sample sizes. Moreover, in research or development, it is important to track the validation accuracy of a model as it trains. Although it is not supported with post-statistics of BN, we can use a simple engineering trick in training US-Nets: always tracking BN statistics of the model at largest and smallest width during training.

\section{Improved Training Techniques}
In this section, we describe our training algorithm for US-Nets from bottom to top. We first introduce motivations and details of \textit{the sandwich rule} and \textit{inplace distillation}, and then present the overall algorithm for training universally slimmable networks.

\subsection{The Sandwich Rule} \label{secs:sandwich}
To train a US-Net, a natural solution is to accumulate or average losses sampled from different sub-networks. For example, in each training iteration we randomly sample \(n\) widths in the range of {\([0.25, 1.0]\times\)} and apply gradients back-propagated from accumulated loss. Taking a step further, the bounded inequality in Equation~\ref{eqs:rethink} tells that in a US-Net, performances at all widths are bounded by performance of the model at smallest width {\(0.25\times\)} and largest width {\(1.0\times\)}. In other words, optimizing performance lower bound and upper bound can implicitly optimize all sub-networks in a US-Net. Thus, we propose \textit{the sandwich rule} that in each iteration we train the model at smallest width, largest width and (\(n-2\)) random widths, instead of \(n\) random widths. We employ this rule and show better convergence behavior and overall performance in experiments. %

\textit{The sandwich rule} brings two additional benefits. First, as mentioned in Section~\ref{secs:post_statistics}, by training smallest width and largest width, we can explicitly track the validation accuracy of a model as it trains, which also indicates the performance lower bound and upper bound of a US-Net. Second, training the largest width is also important and necessary for our next training technique: \textit{inplace distillation}.

\subsection{Inplace Distillation} \label{secs:distill}
The essential idea behind \textit{inplace distillation} is to transfer knowledge inside a single US-Net from full-network to sub-networks inplace in each training iteration. It is motivated by two-step knowledge distilling~\cite{hinton2015distilling} where a large model is trained first, then its learned knowledge is transferred to a small model by training with predicted class soft-probabilities. In US-Nets, by \textit{the sandwich rule} we train the model at largest width, smallest width and other randomly sampled widths all together in each iteration. Remarkably, this training scheme naturally supports inplace knowledge distillation: we can directly use the predicted label of the model at the largest width as the training label for other widths, while for the largest width we use ground truth.

The proposed \textit{inplace distillation} is simple, efficient, and general. In contrast to two-step knowledge distillation~\cite{hinton2015distilling}, \textit{inplace distillation} is single-shot: it can be implemented inplace in training without additional computation or memory cost. And it is generally applicable to all our tasks including image classification, image super-resolution and deep reinforcement learning. For image classification, we use predicted soft-probabilities by largest width with cross entropy as objective function. In image super-resolution, predicted high-resolution patches are used as labels with either \(\ell_1\) or \(\ell_2\) as training objective. For deep reinforcement learning we take proximal policy optimization algorithm (Actor-Critic)~\cite{schulman2017proximal} as an example. To distill, we run the policy predicted by the model at largest width as roll-outs for training other widths.

In practice, it is important to stop gradients of label tensor predicted by the largest width, which means that the loss of a sub-network will never back-propagate through the computation graph of the full-network. Also, the predicted label is directly computed in training mode if it has batch normalization. It works well and saves additional forward cost of inference mode. We tried to combine both ground truth label and predicted label as training label for sub-networks, using either constant balance of two losses or decaying balance, but the results are worse.

\subsection{Training Universally Slimmable Networks}
Equipped with \textit{the sandwich rule} and \textit{inplace distillation}, the overall algorithm for training US-Nets is revealed in Algorithm~\ref{algos:algo}. For simplicity, calculating post-statistics of BN using Equation~\ref{eqs:exact_averages} is not included. It is noteworthy that: (1) The algorithm is general for different tasks and networks. (2) The GPU memory cost is the same as training individual networks thus we can use the same batch size. (3) In all our experiments, same hyper-parameters are applied. (4) It is relatively simple to implement and we show PyTorch-Style pseudo code as an example in Algorithm~\ref{algos:algo}.

\begin{algorithm}[ht]
\caption{Training universally slimmable network \(M\).}
    \begin{algorithmic}[1]
    \Require{Define \textit{width range}, for example, { \([0.25, 1.0] \times\)}.}
    \Require{Define \textit{n} as number of sampled widths per training iteration, for example, \(n=4\).}
    \State{Initialize training settings of shared network \(M\).}
    \For {\(t = 1, ..., T_{iters}\)}
        \State{Get next mini-batch of data \(x\) and label \(y\).}
        \State{Clear gradients, \(optimizer.zero\_grad()\).}
        \State{Execute full-network, \(y' = M(x)\).}
        \State{Compute loss, \(loss = criterion(y', y)\).}
        \State{Accumulate gradients, \(loss.backward()\).}
        \State{Stop gradients of \(y'\) as label, \(y' = y'.detach()\).}
        \State{Randomly sample (\(n-2\)) widths, as \textit{width samples}.}
        \State{Add smallest width to \textit{width samples}.}
        \For {\textit{width} in \textit{width samples}}
            \State{Execute sub-network at \textit{width}, \(\hat{y} = M'(x)\).}
            \State{Compute loss, \(loss = criterion(\hat{y}, y')\).}
            \State{Accumulate gradients, \(loss.backward()\).}
        \EndFor
        \State{Update weights, \(optimizer.step()\).}
    \EndFor
    \end{algorithmic}
\label{algos:algo}
\end{algorithm}

\section{Experiments}
In this section, we first present experiments on tasks of ImageNet classification, image super-resolution and deep reinforcement learning. Next we provide extensive ablation studies regarding \textit{the sandwich rule} and \textit{inplace distillation}. We further study US-Nets with regard to size of samples for BN post-statistics \(s\), width lower bound \(k_0\), width divisor \(d\) and number of sampled widths per training iteration \(n\). In all tables and figures, we use I-Net to denote individually trained models at different widths, S-Net to denote 4-switch slimmable networks~\cite{yu2018slimmable} and US-Net to denote our proposed universally slimmable networks.

\subsection{Main Results}
\textbf{ImageNet Classification.} We experiment with the ImageNet~\cite{deng2009imagenet} classification dataset with \num{1000} classes. Two representative mobile network architectures, MobileNet v1~\cite{howard2017mobilenets} and MobileNet v2~\cite{sandler2018inverted}, are evaluated. Note that MobileNet v1 is a non-residual network, while MobileNet v2 is a residual network.

\begin{table}[ht]
\centering
\caption{Results (top-1 error) on ImageNet classification of I-Net~\cite{howard2017mobilenets, sandler2018inverted}, S-Net~\cite{yu2018slimmable} and US-Net, given same width configurations and FLOPs.}
\begin{tabular}{@{}l@{}l l c c c@{}} \toprule
Network \ & Width & FLOPs & I-Net & S-Net & US-Net\\
\midrule
\multirow{5}{*}{\scriptsize MobileNet v1 \ } & \(1.0 \times\) & 569M & 29.1 & 28.5 \textsubscript{(0.6)} & 28.2 \textsubscript{(0.9)}\\
&\(0.75 \times\) & 317M & 31.6 & 30.5 \textsubscript{(1.1)} & 30.5 \textsubscript{(1.1)}\\
&\(0.5 \times\)  & 150M & 36.7 & 35.2 \textsubscript{(1.5)} & 35.8 \textsubscript{(0.9)}\\
&\(0.25 \times\) & 41M & 50.2 & 46.9 \textsubscript{(3.3)} & 44.3 \textsubscript{(5.9)}\\
\cmidrule{2-6}
& AVG & 269M & 36.9 & 35.3 \textsubscript{(1.6)} & \textbf{34.7 \textsubscript{(2.2)}} \\
\midrule
\multirow{5}{*}{\scriptsize MobileNet v2\ } &\(1.0 \times\) & 301M & 28.2 & 29.5 \textsubscript{(-1.3)} & 28.5 \textsubscript{(-0.3)}\\
&\(0.75 \times\) & 209M & 30.2 & 31.1 \textsubscript{(-0.9)} & 30.3 \textsubscript{(-0.1)}\\
&\(0.5 \times\)  & 97M & 34.6 & 35.6 \textsubscript{(-1.0)} & 35.0 \textsubscript{(-0.4)}\\
&\(0.35 \times\) & 59M & 39.7 & 40.3 \textsubscript{(-0.6)} & 37.8 \textsubscript{(1.9)}\\
\cmidrule{2-6}
& AVG & 167M & 33.2 & 34.1 \textsubscript{(-0.9)} & \textbf{32.9 \textsubscript{(0.3)}} \\
\bottomrule
\end{tabular}
\label{tabs:main_results}
\end{table}

\begin{figure}[ht]
\centering
\includegraphics[width=0.9\linewidth]{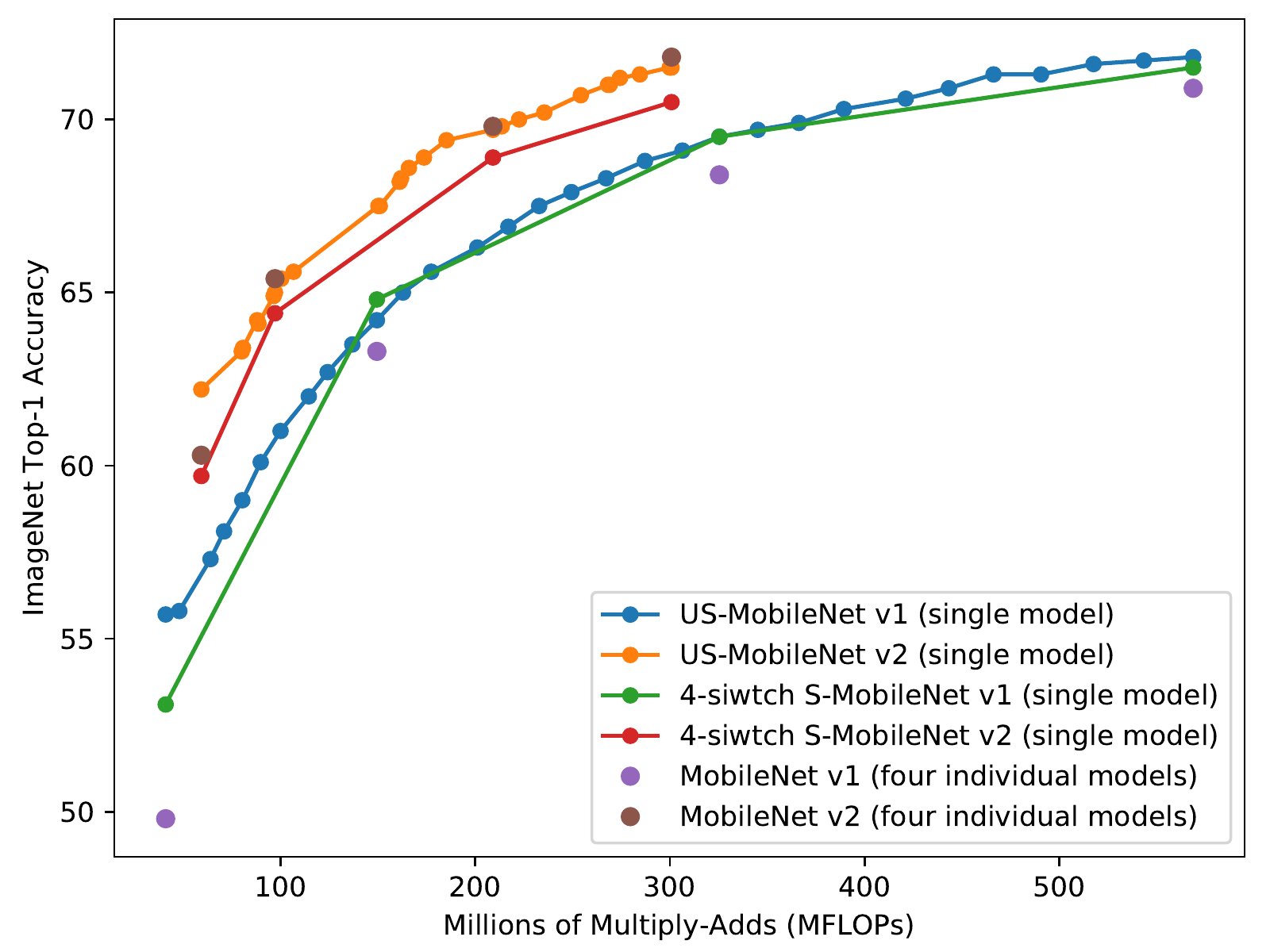}
\caption{FLOPs-Accuracy spectrum of US-MobileNet v1 and US-MobileNet v2, compared with I-Net~\cite{howard2017mobilenets, sandler2018inverted} and S-Net~\cite{yu2018slimmable}.}
\label{figs:main_results}
\end{figure}

We use default training and testing settings in~\cite{howard2017mobilenets, sandler2018inverted} except: (1) We only train US-Nets for \num{250} epochs instead of \num{480} epochs for fast experimentation. (2) We use stochastic gradient descent as the optimizer instead of the \textit{RMSProp}. (3) We decrease learning rate linearly from \(0.5\) to \(0\) with batch size \num{1024} on \num{8} GPUs. We always report results with the model of final training epoch. To be fair, we use \(n=4\) for training US-Nets following Algorithm~\ref{algos:algo}.

We first show numerical results in Table~\ref{tabs:main_results}. Compared with individual models and 4-switch slimmable networks~\cite{yu2018slimmable}, US-Nets have better classification accuracy on average. In Figure~\ref{figs:main_results}, we show FLOPs-Accuracy spectrum of US-MobileNet v1 at widths of {\([.25:.025:1.0]\times\)} and US-MobileNet v2 at widths of { \([.35:.025:1.0]\times\)}.

\textbf{Image Super-Resolution.} We experiment with DIV2K dataset~\cite{timofte2017ntire} which contains \num{800} training and \num{100} validation 2K-resolution images, on the task of bicubic \(\times 2\) image super-resolution. The network WDSR~\cite{yu2018wide} is evaluated. Note that WDSR network has no batch normalization layer~\cite{ioffe2015batch}, instead weight normalization~\cite{salimans2016weight} is used, which requires no further modification in US-Nets. We first individually train two models at width \(n=32\) and width \(n=64\) with 8 residual blocks. We then train US-Nets that can execute at any width in \([32, 64]\), either with or without proposed \textit{inplace distillation} in Section~\ref{secs:distill}.

\begin{figure}[ht]
\centering
\includegraphics[width=0.9\linewidth]{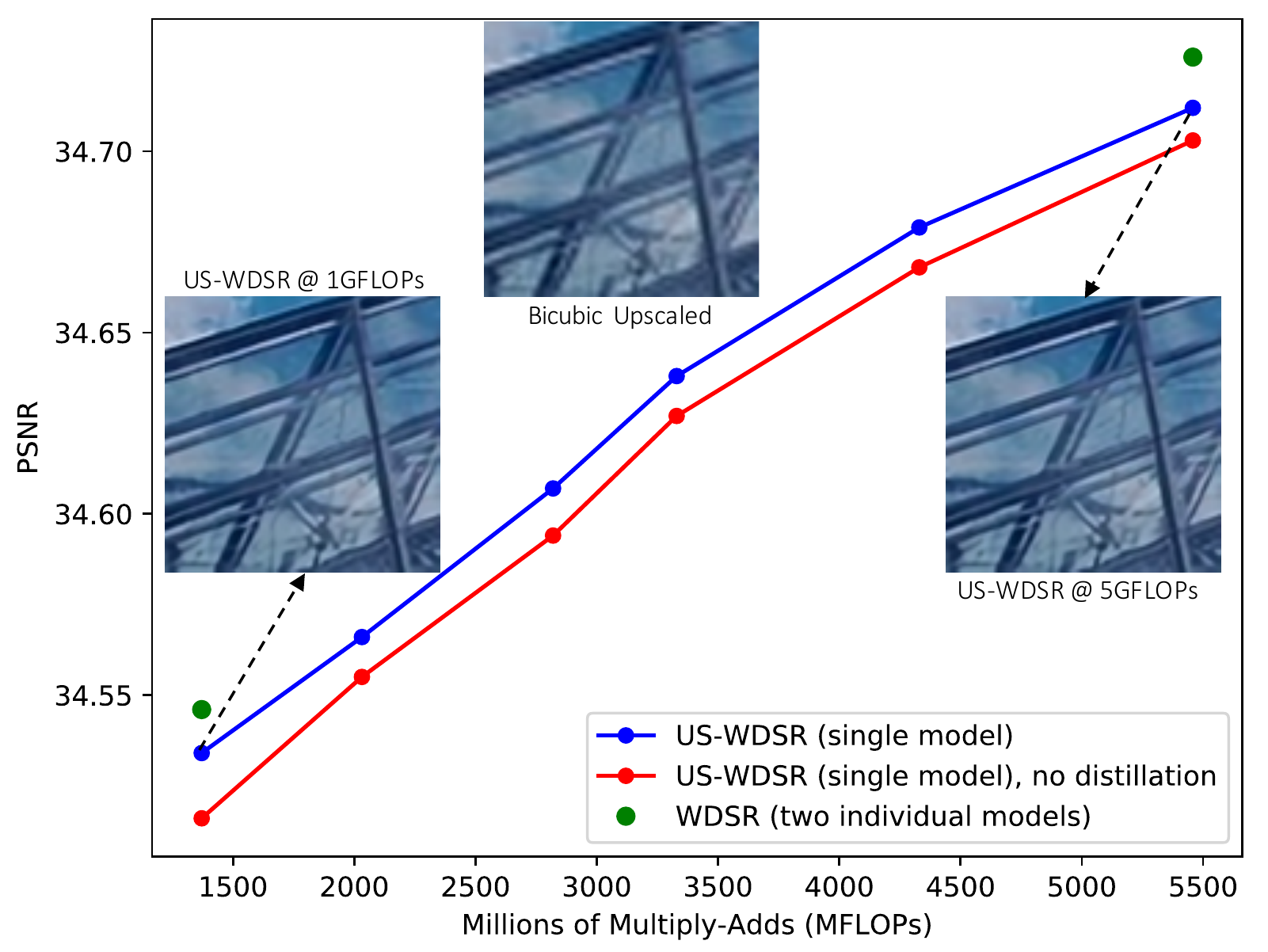}
\caption{FLOPs-PSNR spectrum of US-WDSR and super-resolved high-resolution images under different computations. FLOPs are calculated using input size \(48\times48\).}
\label{figs:image_sr}
\end{figure}

The results are shown in Figure~\ref{figs:image_sr}. US-WDSR have slightly worse performance than individually trained models (but only \(0.01\) lower PSNR). The US-WDSR trained without \textit{inplace distillation} has slightly worse performance. It is noteworthy that we use default hyper-parameters optimized for individual models, which may not be optimal for our slimmable models (e.g., learning rate, initialization, weight decay, etc).

\begin{figure*}[ht]
\centering
\vspace*{-8mm}
\includegraphics[width=\linewidth]{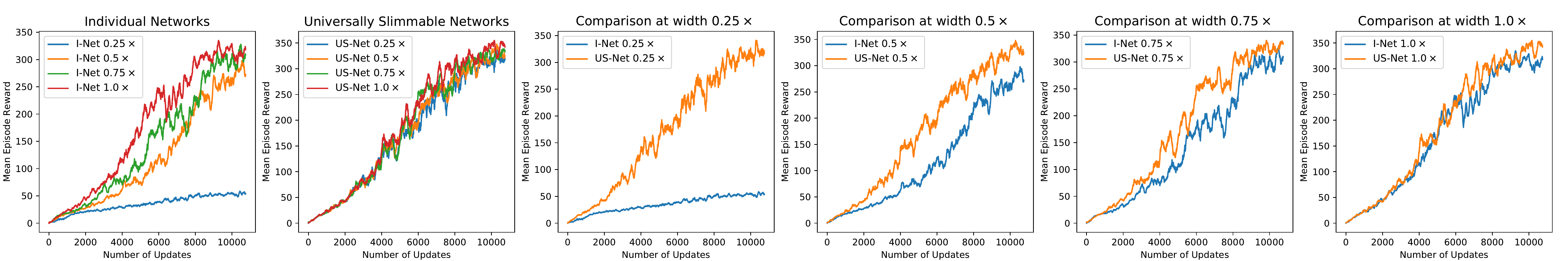}
\caption{Mean Episode Reward with US-Net and I-Net based on actor-critic style PPO~\cite{schulman2017proximal}. Curves are not smoothed.}
\label{figs:deep_rl}
\vspace*{-2mm}
\end{figure*}

\textbf{Deep Reinforcement Learning.} We experiment with Atari game \textit{BreakoutNoFrameskip-v4}~\cite{1606.01540} using Actor-Critic proximal policy optimization algorithm~\cite{schulman2017proximal}. Following baseline models~\cite{schulman2017proximal}, we stack three convolutions with base channel number as \num{32}, \num{64}
, \num{32}, kernel size as \num{8}, \num{4}, \num{3}, stride as \num{4}, \num{2}, \num{1}, and a fully-connected layer with \num{512} output features. The output is shared for both actor (with an additional fully-connected layer to number of actions) and critic (with an additional fully-connected layer to \num{1}). Note that the network has no batch normalization layer.

We first individually train the model at different widths of { \([0.25, 0.5, 0.75, 1.0]\times\)}. Then a US-Net is trained with \textit{inplace distillation} following Section~\ref{secs:distill} and Algorithm~\ref{algos:algo}. The performances are shown in Figure~\ref{figs:deep_rl}. From left to right, we show individually trained models, universally slimmable models (four corresponding widths are shown for comparison), and performance comparison between I-Net and US-Net at widths of { \([0.25, 0.5, 0.75, 1.0]\times\)}. The curves show that the US-Net consistently outperforms four individually trained networks in the task of deep reinforcement learning.

We note that we include the Atari game example mainly to illustrate that our slimmable training is also applicable to CNNs for RL. We believe it is important because in more challenging RL solutions, for example \textit{AlphaGo}~\cite{silver2017mastering} and \textit{AlphaStar}~\cite{alphastar}, the inference latency and adaptive computation ability will be critical.

\subsection{Ablation Study}

\begin{table}[t]
\centering
\caption{Results on ImageNet classification with different \textbf{width sampling rules} during training. We denote {\scriptsize\(\mathit{min}\)} as smallest width, {\scriptsize\(\mathit{max}\)} as largest width, {\scriptsize\(\mathit{random}\)} as randomly sampled widths.}
\small
\begin{tabular}{@{}l c c c c c@{}} \toprule
Sampling Rule & {\small \(0.25 \times\)} & {\small \(0.5 \times\)} & {\small \(0.75 \times\)} & {\small \(1.0 \times\)} & AVG \\
\midrule
{\scriptsize\(3\ \mathit{random}\)} & 55.9 & 35.8 & 31.0 & 30.1 & 38.20\\
{\scriptsize\(\mathit{min+}2\ \mathit{random}\)} & 46.2 & 37.2 & 32.2 & 31.3 & 36.73\\
{\scriptsize\(\mathit{max+}2\ \mathit{random}\)} & 58.4 & 37.0 & 31.1 & 28.3 & 38.70\\
{\scriptsize\(\mathit{min+}1\ \mathit{random}\mathit{+max}\)} & 46.6 & 38.6 & 32.4 & 28.2 & \textbf{36.45}\\
\bottomrule
\end{tabular}
\label{tabs:the_sandwich_rule}
\vspace{-2mm}
\end{table}

\textbf{The Sandwich Rule.} We study the effectiveness of \textit{the sandwich rule} by ablation experiments. We train four models of US-MobileNet v1 with \(n=3\) using different width sampling rules: \(n\) randomly sampled widths, (\(n-1\)) randomly sampled widths plus the smallest width, (\(n-1\)) randomly sampled widths plus the largest width, and (\(n-2\)) randomly sampled widths plus both the smallest and largest width. Results are shown in Table~\ref{tabs:the_sandwich_rule}. The US-Net trained with \textit{the sandwich rule} has better performance on average, with good accuracy at both smallest width and largest width. Moreover, training the model at smallest width is more important than training the model at largest width as shown in the 2nd row and 3rd row of Table~\ref{tabs:the_sandwich_rule}, which suggests the importance of width lower bound \(k_0\). \textit{Inplace distillation} is not used in all these experiments since it is not applicable to width sampling rules excluding largest width.

\begin{figure}[ht]
\centering
\includegraphics[width=0.9\linewidth]{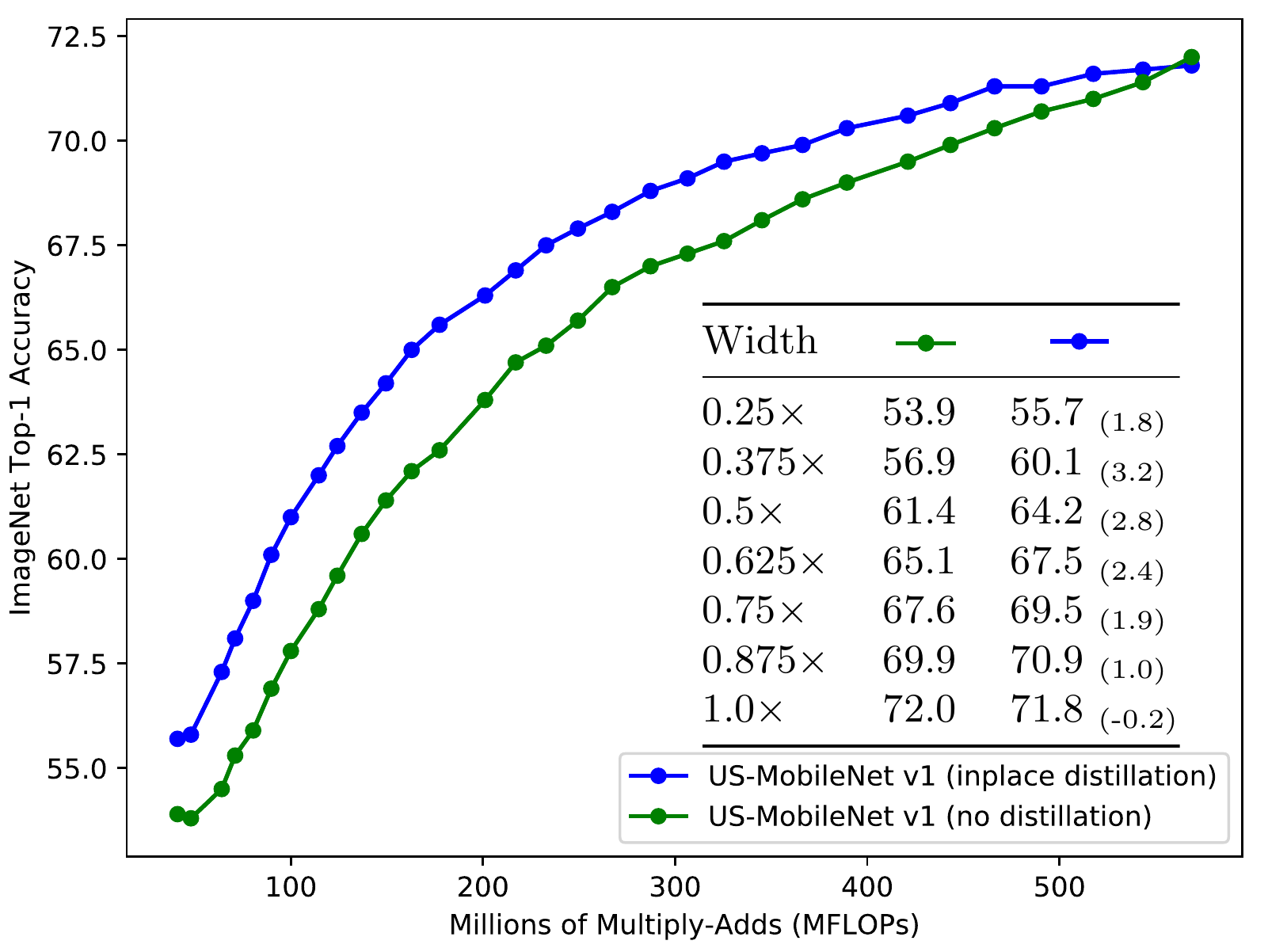}
\caption{FLOPs-Accuracy spectrum of two US-MobileNet v1 models trained either with or without \textit{inplace distillation}.}
\label{figs:distillation}
\vspace{-2mm}
\end{figure}

\textbf{Inplace Distillation.} Next we study the effectiveness of proposed \textit{inplace distillation} mainly on ImageNet classification. The results of image super-resolution (both with and without \textit{inplace distillation}) and deep reinforcement learning (with \textit{inplace distillation}) are already shown in Figure~\ref{figs:image_sr} and Figure~\ref{figs:deep_rl}. We use the same settings to train two US-MobileNet v1 models either with or without \textit{inplace distillation}, and show the comparison in Figure~\ref{figs:distillation}. \textit{Inplace distillation} significantly improves overall performance at no cost. We suppose it could be an essential component for training slimmable networks.

\begin{table}[t]
\centering
\caption{Performance comparison (top-1 error) of different methods for calculating \textbf{post-statistics of batch normalization}. We use either moving (Equation~\ref{eqs:moving_averages}) or exact (Equation~\ref{eqs:exact_averages}) averages.}
\small
\begin{tabular}{@{}l l c c c c c@{}} \toprule
Size of Samples & Average & {\small \(0.25 \times\)} & {\small \(0.5 \times\)} & {\small \(0.75 \times\)} & {\small \(1.0 \times\)}\\
\midrule
1.28M & Moving & 44.4 & 35.8 & 30.6 & 28.2\\
1.28M & Exact & \textbf{44.3} & 35.8 & \textbf{30.5} & 28.2\\
1k & Exact & 44.4 & 35.8 & 30.6 & 28.2\\
2k & Exact & \textbf{44.3} & 35.8 & \textbf{30.5} & 28.2\\
\bottomrule
\end{tabular}
\label{tabs:batch_norm}
\vspace{-2mm}
\end{table}

\textbf{Post-Statistics of Batch Normalization.} We further study  post-statistics for batch normalization in US-Nets. We update BN statistics after training US-MobileNet v1 when all weights are fixed. We then compute BN statistics using four methods: moving average over entire training set, exact average over entire training set, exact average over randomly sampled \(1k\) training subset, and exact average over randomly sampled \(2k\) training subset. Table~\ref{tabs:batch_norm} shows that exact averaging has slightly better performance and a small subset produces equally accurate BN statistics. It indicates that calculating post-statistics of BN can be very fast.

\begin{figure}[ht]
\centering
\includegraphics[width=0.9\linewidth]{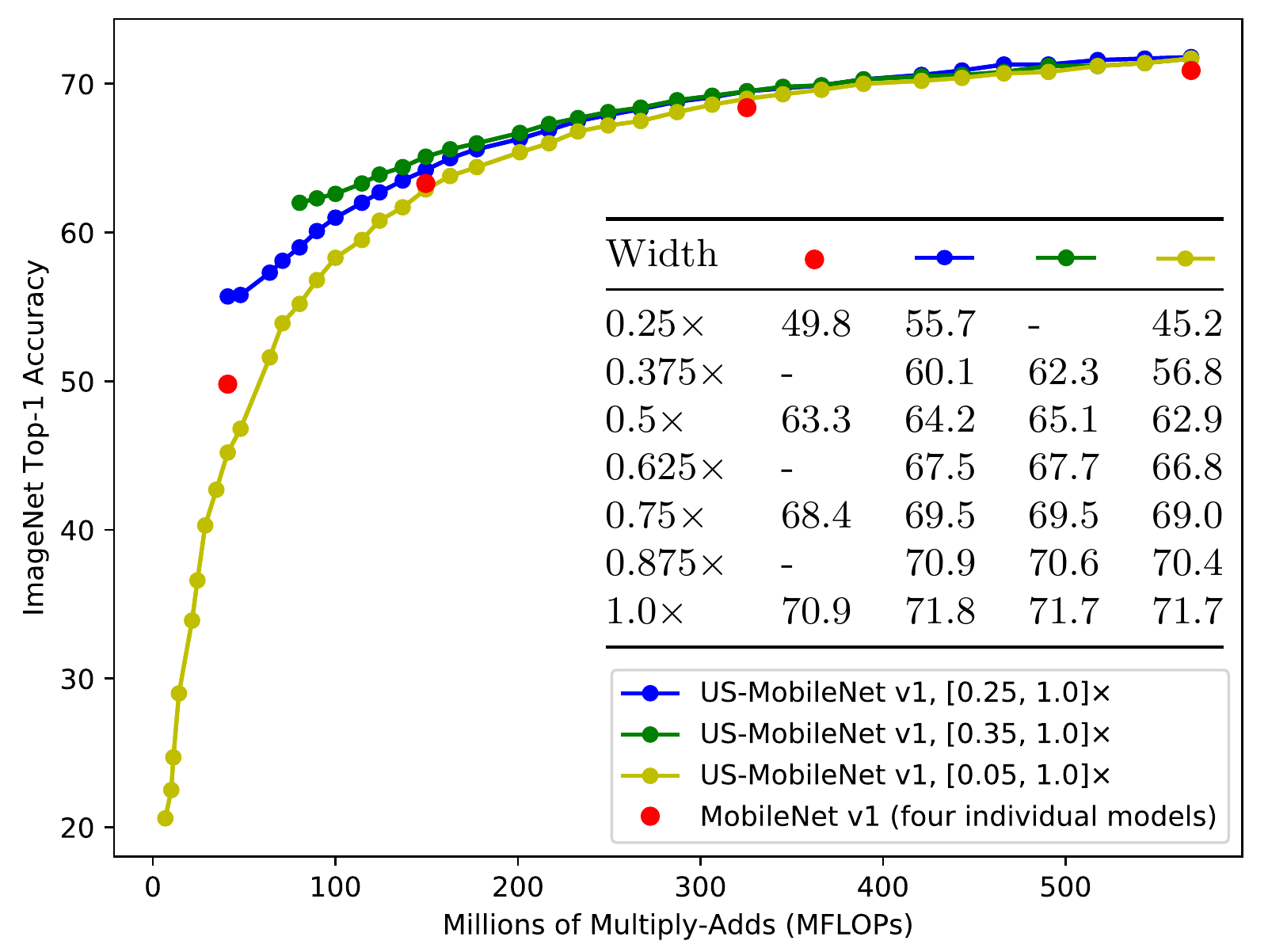}
\caption{FLOPs-Accuracy spectrum of three US-MobileNet v1 models with different \textbf{width lower bounds}.}
\label{figs:lower_bound}
\vspace{-2mm}
\end{figure}

\textbf{Width Lower Bound \(k_0\).} Width lower bound \(k_0\) is of central importance in the bounded Equation~\ref{eqs:rethink}. Although it is usually enough to adjust a model between width { \(0.25\times\)} and { \(1.0\times\)}, we are interested in how the width lower bound affects overall performance. We train three US-MobileNet v1 models with different width lower bounds \(k_0\) as {\(0.25\times\)}, {\(0.35\times\)}, {\(0.05\times\)} and show results in Figure~\ref{figs:lower_bound}. It reveals that the performance of a US-Net is grounded on its width lower bound, as suggested in our analysis in Section~\ref{secs:rethink}.

\begin{figure}[ht]
\centering
\includegraphics[width=0.9\linewidth]{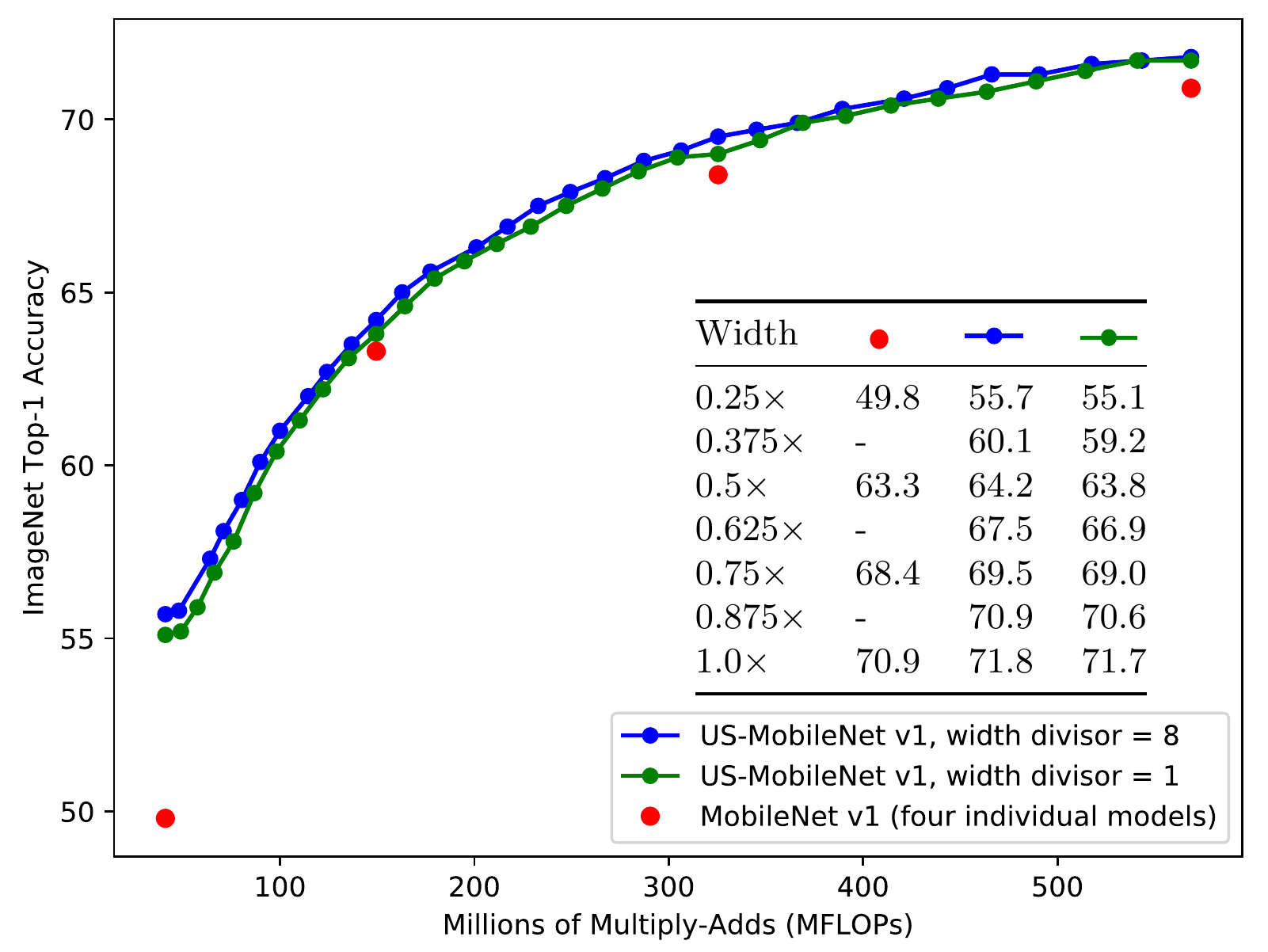}
\caption{FLOPs-Accuracy spectrum of two US-MobileNet v1 models with different \textbf{width divisors}.}
\label{figs:width_divisor}
\vspace{-2mm}
\end{figure}

\textbf{Width Divisor \(d\).} Width divisor is introduced in MobileNets~\cite{howard2017mobilenets, sandler2018inverted} to floor the channel number approximately as \(\lfloor nr/d \rfloor * d\), where \(n\) is base channel number, \(r\) is width multiplier, \(d\) is width divisor\footnote{Details are in hyperlink \href{https://github.com/tensorflow/models/blob/0344c5503ee55e24f0de7f37336a6e08f10976fd/research/slim/nets/mobilenet/mobilenet.py\#L62-L69}{TensorFlow Models} (PDF required).}. To exactly match FLOPs of MobileNets and have a fair comparison, by default we follow MobileNets and set width divisor \(d=8\). This results in the minimal adjustable channel number as \(8\) instead of \(1\), and slightly benefits overall performance, as shown in Figure~\ref{figs:width_divisor}. In practice, with \(d=8\) the US-Nets already provide enough adjustable widths. Also in many hardware systems, matrix multiplication with size dividable by \(d=8, 16, ...,\) may be as fast as a smaller size due to alignment of processing unit (\eg, warp size in GPU is 32).

\begin{figure}[ht]
\centering
\includegraphics[width=0.9\linewidth]{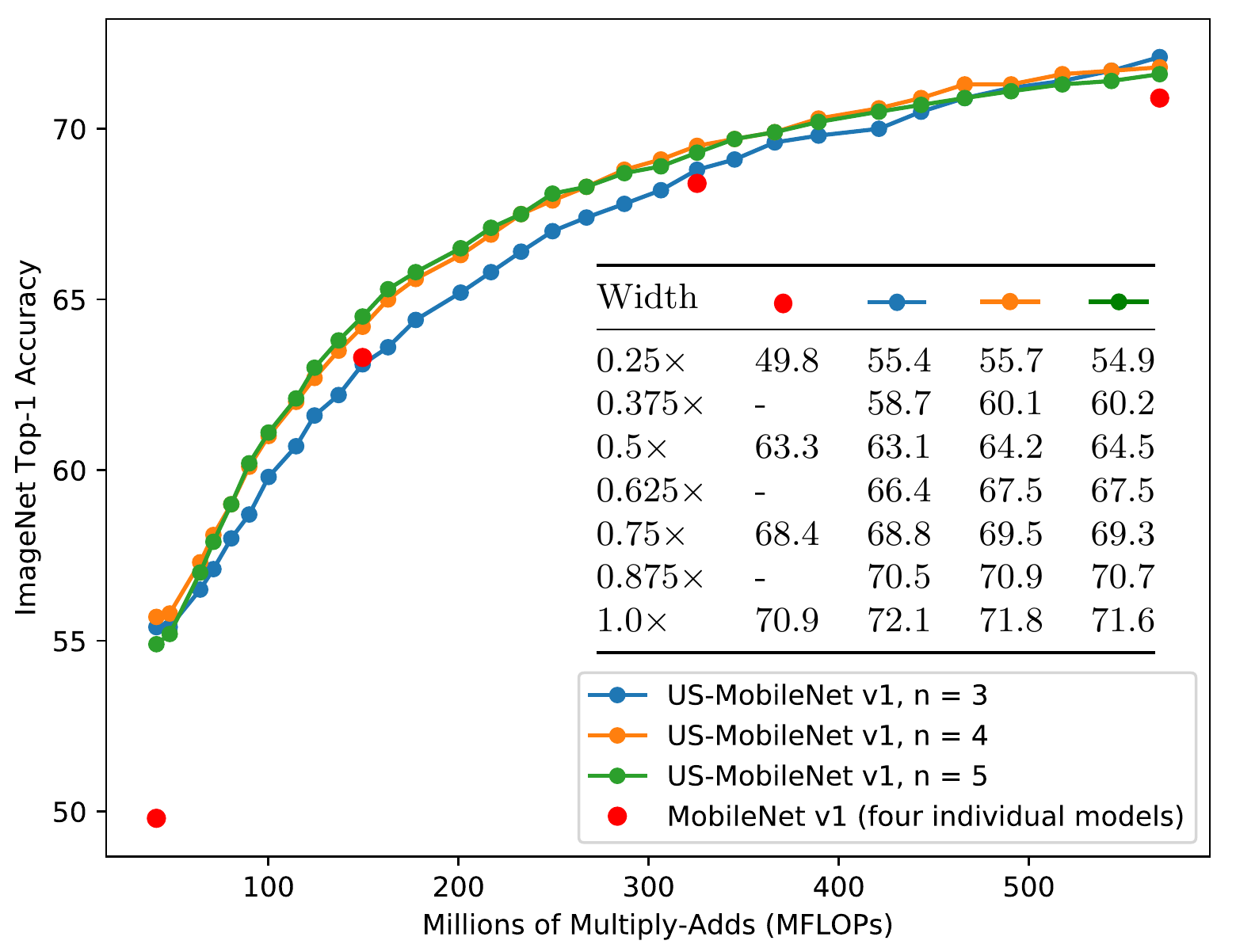}
\caption{FLOPs-Accuracy spectrum of two US-MobileNet v1 trained with different \textbf{numbers of sampled widths per iteration}.}
\label{figs:num}
\vspace{-2mm}
\end{figure}

\textbf{Number of Sampled Widths Per Iteration \(n\).} Finally we study the number of sampled widths per training iteration. It is important because larger \(n\) leads to more training time. We train three US-MobileNet v1 models with \(n\) equal to \(3\), \(4\) or \(5\). Figure~\ref{figs:num} shows that the model trained with \(n=4\) has better performance than the one with \(n=3\), while \(n=4\) and \(n=5\) achieve very similar performances. By default, in all our experiments we use \(n=4\).

\section{Discussion} \label{secs:discuss}
We mainly discuss three topics in this section, with detailed results shown in the supplementary materials.

First, for all trained US-Nets so far, the width ratio is uniformly applied to all layers. Can we train a nonuniform US-Net where each layer can independently adjust its own ratio using our proposed methods? This requirement is especially important for related tasks like network slimming. Our answer is YES and we show a simple demonstration on how the nonuniform US-Net can help in network slimming.

Second, perhaps the question is naive, but are deep neural networks naturally slimmable? The answer is NO, a naively trained model fails to run at different widths even if their BN statistics are calibrated.

Third, in slimmable networks~\cite{yu2018slimmable}, private scale and bias are used as conditional parameters for each sub-network, which brings performance gain slightly. In US-Nets, by default we share scale and bias. We also propose an option that mimics conditional parameters: averaging the output by the number of input channels.

{\small
\bibliographystyle{ieee_fullname}
\bibliography{egbib}
}

\newpage
\appendix
\section{Discussion} \label{secs:discuss}
In this section, we mainly discuss three topics with some experimental results.

\textbf{Nonuniform Universally Slimmable Networks.} For all trained US-Nets so far, the width ratio is uniformly applied to all layers (\eg, MobileNet { \(0.25 \times\)} means width in all layers are scaled by \(0.25\)). Can we train a nonuniform US-Net where each layer can independently adjust its own ratio using our proposed methods? This requirement is especially important for related tasks like network slimming. Our answer is YES and we show a simple demonstration on how the nonuniform US-Net can help in network slimming.

\begin{figure}[ht]
\centering
\includegraphics[width=\linewidth]{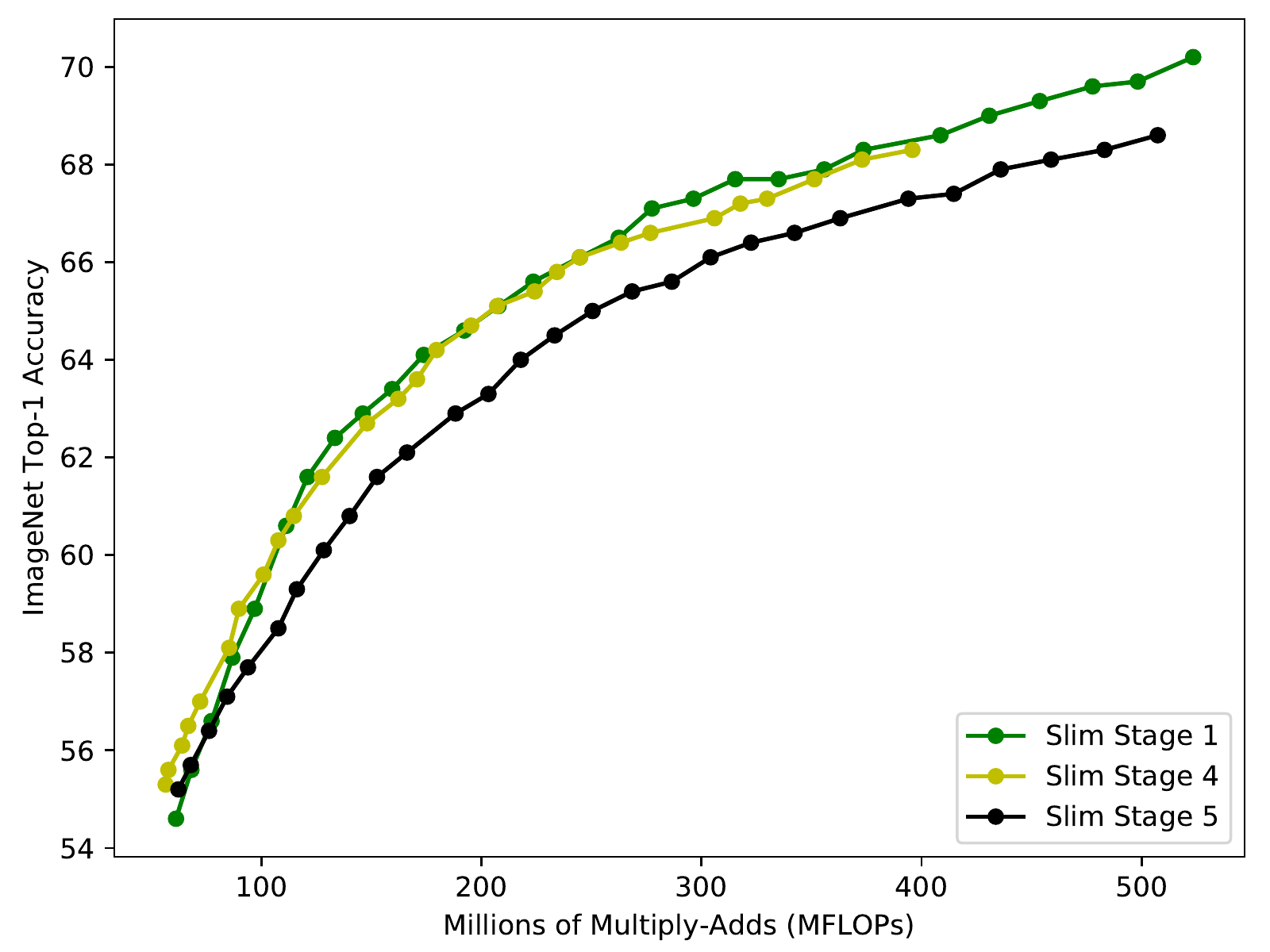}
\caption{FLOPs-Accuracy spectrum of nonuniform US-MobileNet v1 tested with different slimming strategies. Note that each layer can adjust its own width ratio. The result suggests that \textit{slimming the stage 5 of MobileNet v1 is not a good choice}.}
\label{figs:nonuniform}
\end{figure}

In this demonstration, we first train a nonuniform US-MobileNet v1. The architecture of MobileNet v1 has 5 resolution stages with base channel number as 64, 128, 256, 512, 1024 in each stage. After training, we apply an additional width ratio \(0.6\) to one of five stages and get five models. Along with global width ratio, we can draw their FLOPs-Accuracy spectrum in Figure~\ref{figs:nonuniform}. For simplicity we only show performances of slimming stage 1, 4 and 5. Slimming stage 2 and 3 have curves close to that of slimming stage 1, while slimming stage 1 achieves the best results. Figure~\ref{figs:nonuniform} shows that the stage 5 of MobileNet v1 may require more channels because slimming stage 5 has worst accuracy under same FLOPs. The result suggests \textit{slimming the stage 5 of MobileNet v1 is not a good choice}. It further implicitly indicates that the stage 5 of MobileNet v1 network architecture needs a larger base channel number.

\begin{figure}[ht]
\centering
\includegraphics[width=\linewidth]{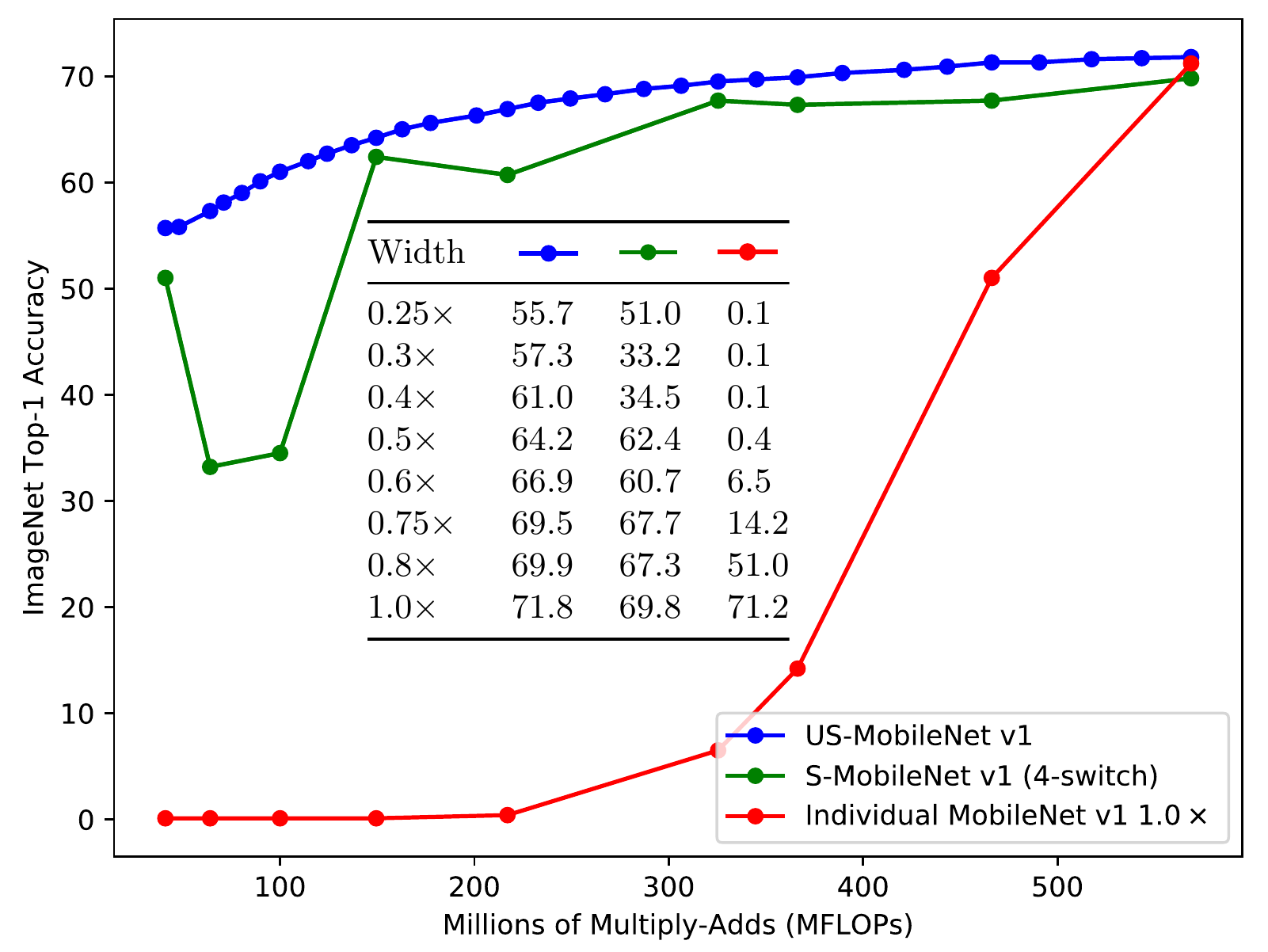}
\caption{FLOPs-Accuracy spectrum of US-MobileNet v1, 4-switch S-MobileNet v1 and individual MobileNet v1 { \(1.0\times\)} tested on different widths after BN calibration. The results suggest that deep neural networks are not naturally slimmable.}
\label{figs:naturally_slimmable}
\end{figure}

\textbf{Naturally Slimmable?} Perhaps the question is naive, but are deep neural networks naturally slimmable? We have proposed training methods and improved techniques for universally slimmable networks, yet we have not presented any result if we directly evaluate a trained neural network at arbitrary width either with naive training algorithm or slimmable training algorithm in~\cite{yu2018slimmable}. If we can calibrate post-statistics of BN in these trained models (instead of using our proposed US-Nets training algorithm), do they have good performances? The answer is NO, both naively trained models and slimmable models~\cite{yu2018slimmable} have very low accuracy at arbitrary widths even if their BN statistics are calibrated.

In Figure~\ref{figs:naturally_slimmable}, we show results of a US-MobileNet v1, 4-switch S-MobileNet v1 { \([0.25, 0.5, 0.75, 1.0]\times\)} and individually trained MobileNet v1 { \(1.0\times\)}. For individually trained MobileNet v1 { \(1.0\times\)}, it achieves good accuracy at width { \(1.0\times\)}, but fails on other widths especially when its computation is below 200 MFLOPs. For 4-switch S-MobileNet v1 { \([0.25, 0.5, 0.75, 1.0]\times\)}, it achieves good accuracy at widths in { \([0.25, 0.5, 0.75, 1.0]\times\)}, but fails on other widths that are not included in training. Our proposed US-MobileNet v1 achieves good accuracy at any width in the range from 40 MFLOPs to 570 MFLOPs consistently.

\textbf{Averaging Output by Input Channel Numbers.} In slimmable networks~\cite{yu2018slimmable}, private scale and bias \(\gamma\), \(\beta\) are used as conditional parameters for each sub-network, which brings slight performance gain. These parameters comes for free because after training, they can be merged as \(y' =  \gamma'y + \beta', \gamma' = \frac{\gamma}{\sqrt{\sigma^2 + \epsilon}}, \beta' = \beta - \gamma'\mu\).

In US-Nets, by default we share scale and bias. Additionally we propose an option that mimics conditional parameters: averaging the output by the number of input channels. It also brings slight performance gain as shown in Table~\ref{tabs:averaged_output}. In this way, to some extent the \textit{feature aggregation} can be viewed as \textit{feature ensemble} in each layer.

\begin{table}[h]
\centering
\caption{Performance comparison (top-1 error) of our default model (US-MobileNet v1) and model trained with \textbf{output averaging} (US-MobileNet v1 +).}
\small
\begin{tabular}{@{}l l l l l l l@{}} \toprule
Name & {\small \(0.25 \times\)} & {\small \(0.5 \times\)} & {\small \(0.75 \times\)} & {\small \(1.0 \times\)} & AVG\\
\midrule
{\scriptsize US-MobileNet v1} & 44.3 & 35.8 & 30.5 & 28.2 & 34.7\\
{\scriptsize US-MobileNet v1 +} & 43.3 & 35.5 & 30.6 & 27.9 & \textbf{34.3} \textsubscript{(0.4)}\\
\bottomrule
\end{tabular}
\label{tabs:averaged_output}
\end{table}

In practice, it is important not to average depthwise convolution, because the actual input to each output channel in depthwise convolution is always single-channel. For networks with batch normalization, the proposed output averaging also come for free since these constants can be merged into BN statistics after training. At runtime when switch to different widths, a switch cost (\eg., fusing new BN to its previous convolution layer) will be applied. But for networks without batch normalization, we should notice that if we do not use output averaging, there is no switch cost. Thus, the proposed output averaging is optional and is not used by default.

\end{document}